\documentclass[lettersize,journal]{IEEEtran}
\usepackage{amsmath,amsfonts}
\usepackage{algorithmic}
\usepackage{algorithm}
\usepackage{array}
\usepackage{textcomp}
\usepackage{stfloats}
\usepackage{url}
\usepackage{verbatim}
\usepackage[caption=false,font=normalsize,labelfont=sf,textfont=sf]{subfig}
\usepackage{cite}
\usepackage{graphicx}    
\graphicspath{{figs/}}    
\usepackage{caption}
\usepackage{times}
\usepackage{booktabs}
\usepackage{multirow}

\begin{document}
\captionsetup[figure]{labelfont={bf},labelformat={default},labelsep=period,name={Fig.}}
\captionsetup[table]{labelfont={},labelformat={default},labelsep=period,name={Table}}
\captionsetup[subfloat]{font=footnotesize,labelfont=rm,textfont=rm}

\title{Learning to Calibrate for Reliable Visual Fire Detection}

\author{Ziqi Zhang\textsuperscript{1,2}, Xiuzhuang Zhou\textsuperscript{1}, and Xiangyang Gong\textsuperscript{2}
\thanks{\textsuperscript{1} Authors are with the School of Intelligent Engineering and Automation, Beijing University of Posts and Telecommunications, Beijing 100876, China(e-mail: zzq75071@bupt.edu.cn)}
\thanks{\textsuperscript{2} Authors are with Information and Digital Management Department, China Petrochemical Corporation, Beijing 100728, China}
}
\maketitle

\begin{abstract}
Fire is characterized by its sudden onset and destructive power, making early fire detection crucial for ensuring human safety and protecting property. With the advancement of deep learning, the application of computer vision in fire detection has significantly improved. However, deep learning models often exhibit a tendency toward \textit{overconfidence}, and most existing works focus primarily on enhancing classification performance, with limited attention given to uncertainty modeling. To address this issue, we propose transforming the Expected Calibration Error (ECE), a metric for measuring uncertainty, into a differentiable ECE loss function. This loss is then combined with the cross-entropy loss to guide the training process of multi-class fire detection models. Additionally, to achieve a good balance between classification accuracy and reliable decision, we introduce a curriculum learning-based approach that dynamically adjusts the weight of the ECE loss during training. Extensive experiments are conducted on two widely used multi-class fire detection datasets, DFAN and EdgeFireSmoke, validating the effectiveness of our uncertainty modeling method.
\end{abstract}

\section{Introduction}
Fire detection involves the identification and confirmation of fire events in a monitored environment using various technical methods, enabling timely interventions to control the fire and minimize associated damage. This technology is widely employed in fire-prone settings, such as chemical plants, construction material companies, and residential areas, and is typically implemented through sensor-based detection, human monitoring, or algorithmic analysis.

Due to the inherent limitations of sensor detection technologies and the low efficiency of manual patrols, these methods often fail to meet the real-time requirements of fire detection. As a result, most of the attempts have shifted towards the application of computer vision, which enables continuous, intelligent fire detection and alarm systems based on video or image data. In \cite{1liang2023}, a multi-feature fusion detection method is proposed to leverage distinct visual characteristics \cite{2jin2023} associated with fire to identify regions in images that match these features, thereby determining the presence of a fire. Additionally, deep learning techniques are employed to automatically learn richer, high-level features from video or image data. Disturbance removal methods \cite{3tao2024} and the design of more discriminative fire features \cite{4he2021} further reduce the impact of complex environmental factors on the model's recognition accuracy. Visual fire detection offers the advantage of directly identifying fire locations, making it particularly effective in high-risk environments, such as areas containing hazardous gases, where traditional detection methods may fall short. In recent years, with advancements in deep learning, visual fire detection has seen substantial improvements, leading to a significant enhancement in recognition accuracy.

\begin{figure}[!htbp]
    \centering
    \subfloat[Red Maple Trees]{\includegraphics[width=1.64in, height=1.2in]{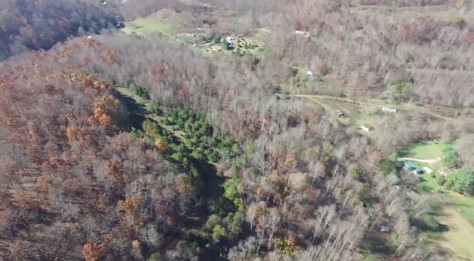}}\hfill
    \subfloat[Flaming Clouds]{\includegraphics[width=1.64in, height=1.2in]{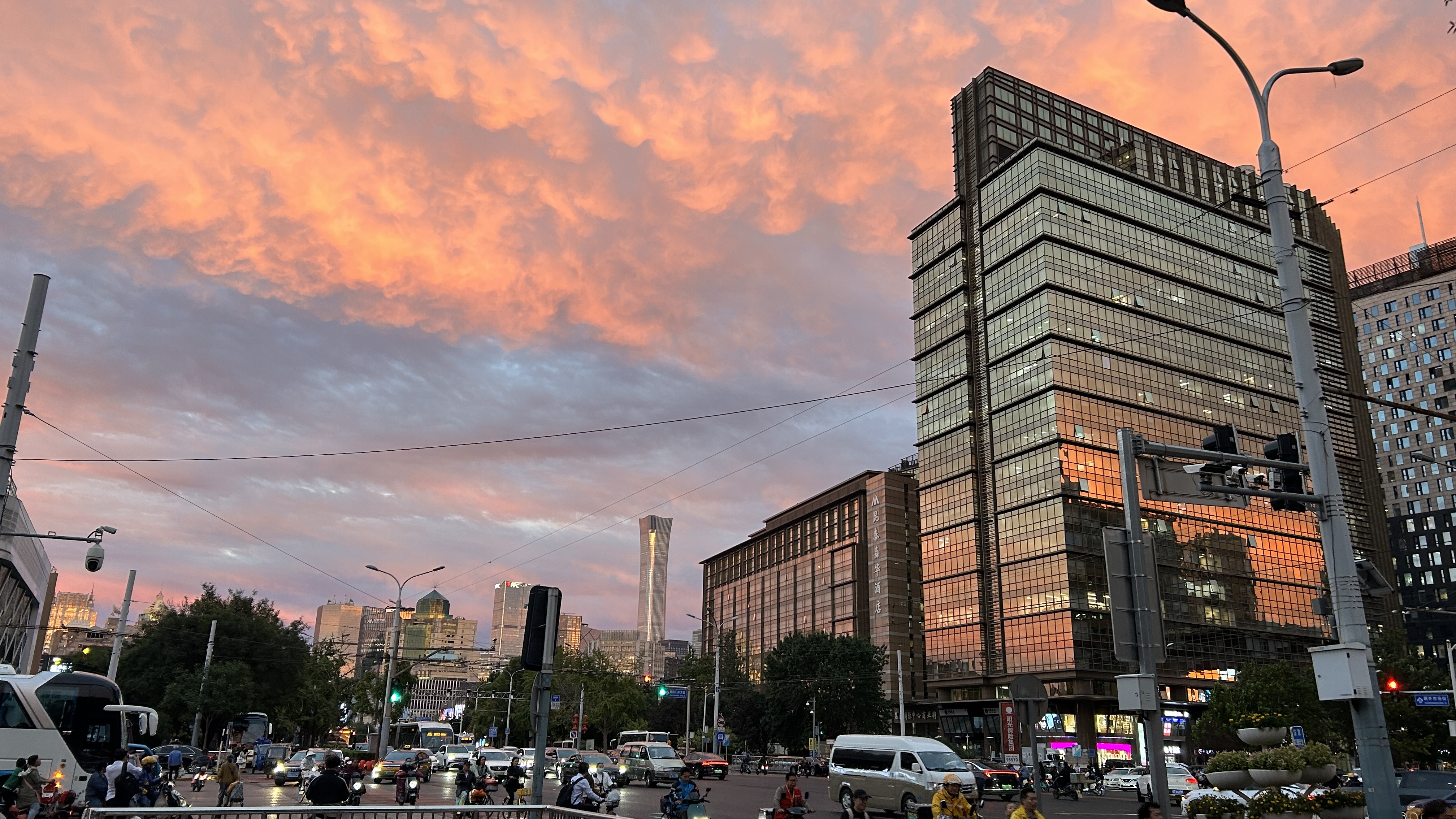}} \\
    \subfloat[Headlights]{\includegraphics[width=1.64in, height=1.2in]{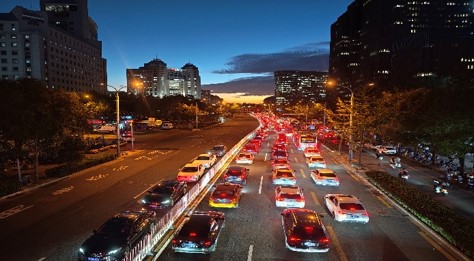}}\hfill
    \subfloat[Water Vapor]{\includegraphics[width=1.64in, height=1.2in]{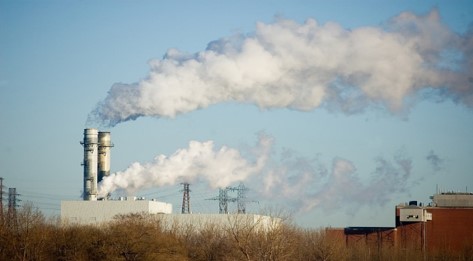}} \\
    \caption{Non-fire images with interfering objects.}
    \label{fig:1}
\end{figure}

Fire detection systems are generally based on classification models that judge whether a fire occurs. In this process, the correctness of the decision largely depends on the accuracy of the predicted probabilities. However, fire detection scenarios are usually quite complex, with many interfering objects in the environment, as shown in Figure \ref{fig:1}, which is highly similar to flames or smoke in terms of color, fluidity, and other features. Additionally, flames and smoke have irregular shapes and varying shades\cite{5QHXB2024}, and the visual characteristics of different objects when burning may also be distinctly different. For example, objects like sulfur and magnesium may exhibit rare colors such as blue-purple when burning, sometimes accompanied with other visual characteristics such as emitting white light. The features of fire images are similar to those of non-fire images, while the features among different fire images can vary greatly, leading to higher uncertainty.

Predictive uncertainty can significantly impact the accuracy of a model's decisions, leading to false positives or false negatives, which may result in severe, irreparable consequences. Therefore, modeling uncertainty is crucial for achieving a more comprehensive understanding of fire detection models and mitigating the risk of over-reliance on inaccurate predictions. Analyzing the sources of uncertainty can also facilitate targeted improvements to the model, such as adjustments to the model architecture or enhancements in data processing, ultimately leading to more reliable decision. However, effectively modeling predictive uncertainty requires substantial computational resources and additional validation/recalibration data \cite{6arlovic2024,7yelleni2024}, which poses a considerable challenge to its practical implementation. Additionally, in classification tasks, achieving an appropriate balance between classification accuracy and reliable decision remains a critical research challenge.

This paper proposes a new method for modeling uncertainty in visual fire detection, by introducing a new loss and calibrating the fire detection model online based on curriculum learning. Our work makes the following contributions:

1) A differentiable ECE Loss is introduced when training the multi-class fire detection models, for modeling the predictive uncertainty in visual fire detection.

2) A method for dynamically adjusting the weight of ECE Loss is designed, inspired by the principles of curriculum learning. This enables the model to progressively transition from simpler to more complex tasks, thereby effectively balancing classification accuracy and reliable decision.

3) Extensive experimental evaluations are conducted on the publicly available datasets DFAN and EdgeFireSmoke for fire detection, demonstrating that the proposed method achieves improved calibration performance without sacrificing the classification accuracy.

\section{Related Work}
Fire detection is designed to identify and confirm the occurrence of a fire in a monitored scene. Currently, fire detection relies mainly on computer vision technology, with works focusing on areas such as distinguishing interfering objects, recognizing complex fire images, and enhancing the efficiency of detection systems.

Fire detection environments are diverse and complex, with many scenarios prone to misidentification. Tao et al. \cite{3tao2024} proposed a triple disturbance removal network for smoke detection, which learned discriminative representations to effectively reduce the false alarm rate caused by disturbances at spatial, temporal, and semantic levels. He et al. \cite{4he2021} introduced a lightweight feature-level and decision-level fusion module, incorporating spatial and channel attention mechanisms to detect small smoke patterns and recognize smoke-like objects. Tao et al. \cite{8tao2023} developed a forest smoke recognition network with pixel-level supervision, featuring a detail difference perception module, an attention feature separation module, and a multi-connection aggregation method, which effectively mitigates the low detection rate and high false alarm rate in complex scenarios. Park et al. \cite{9park2022} proposed a method for generating virtual wildfire images using a Generative Adversarial Network (GAN), annotating them with a weakly supervised image localization module, and performing wildfire detection based on an enhanced YOLOv5s model, significantly reducing false alarms during the detection process.

The visual characteristics of fire vary significantly across different detection scenarios, and the irregular, dynamic shapes of flames and smoke further complicate detection. Li et al. \cite{10li2023} proposed an anchor-free fire recognition algorithm that integrated a multi-scale feature fusion network with a channel attention mechanism, combining loss functions including classification loss, regression loss, and center point loss. This approach enhanced the model's ability to detect irregularly shaped flames and smoke with blurred boundaries. Yuan et al. \cite{11yuan2022} introduced a method that combined a 3D cross-convolutional attention module with count prior embedding, addressing the challenges posed by the semi-transparency and blurred edges of smoke, which often led to reduced detection accuracy. Liang et al. \cite{12laing2024} proposed an anchor-free, structure-based fire detection algorithm, designing the feature extraction network's residual module as a multi-branch structure to capture more expressive flame features. By strengthening feature representation through an improved feature fusion network, this method enhanced the model's ability to detect multi-scale flames, making it suitable for many fire detection scenarios.

Considering the rapid spread of fire, it is crucial not only to improve the accuracy of fire detection but also to enhance the inference speed and deployment efficiency of models. Siddique et al. \cite{13siddique2024} proposed an Internet of Things (IoT)-based federated learning framework for forest fire classification, which distributed computational tasks across multiple nodes. This approach enhanced detection efficiency while safeguarding user privacy and data security. Li et al. \cite{14li2023} introduced a lightweight fire detection model and developed an edge computing system that connected feedback from the edge model to edge gateways and smart devices. This solution addresses the limitations of traditional fire detection systems, which are often too large to be deployed on edge devices. Tian et al. \cite{15tian2024} proposed a fire detection algorithm that strengthened spatial feature extraction and multi-scale feature fusion, incorporating local convolution modules to reduce the size of the backbone network and detection head. This approach achieved high detection accuracy while ensuring real-time performance. Zhang et al. \cite{16zhang2023} presented a flame and smoke detection algorithm that integrated a YOLOv5-ResNet cascade network. By enhancing the YOLOv5 detection network and combining continuous multi-frame detection results with changes in smoke area, the algorithm improved the detection performance of small flame and smoke targets. This approach also effectively eliminated non-flame and non-smoke objects, achieving high accuracy, rapid detection, and low false alarm rate, making it suitable for large-scale industrial applications.

In recent years, the rapid development of deep learning has significantly enhanced the recognition accuracy of visual fire detection. However, models still exhibit a tendency to be \textit{overconfident} in their predictions \cite{17guo2017}. Specifically, for certain samples, the model may produce incorrect classification results while maintaining high confidence in these erroneous predictions. Furthermore, much of current works in visual fire detection focus on improving detection accuracy, with little attention given to uncertainty modeling.

Uncertainty modeling methods have broad applications in the field of computer vision. Ji et al. \cite{18ji2023} were among the first to incorporate uncertainty into the task of image tampering detection. They proposed an uncertainty estimation network that dynamically supervised uncertainty from both the data and the model, using the generated uncertainty map to refine tampering detection outcomes. This approach led to more accurate and reliable detection. In the context of salient object detection, Tian et al. \cite{19tian2023} explored distribution uncertainty, investigating the effectiveness of long-tail learning, single-model uncertainty modeling, and test-time strategies to address the distributional differences between training and testing samples. Yelleni et al. \cite{7yelleni2024} focused on uncertainty in object detection, introducing a method called MC-DropBlock. This approach leveraged the DropBlock technique to model cognitive uncertainty during model training and inference, while using a Gaussian likelihood function to capture accidental uncertainty in the data. Their method significantly enhanced the generalization ability of object detection models.

This paper introduces a new uncertainty modeling method to the visual fire detection by integrating an uncertainty-aware loss with cross-entropy loss and training the model based on curriculum learning. The proposed method demonstrates improved calibration performance in multi-class fire detection tasks, enhancing the reliability of the model's decisions.

\section{Method}
Modeling uncertainty in visual fire detection involves two key aspects: uncertainty calibration and uncertainty measurement. Calibration refers to the process of adjusting a model so that its predicted confidence aligns with its actual classification accuracy, thereby reducing predictive uncertainty. In contrast, uncertainty measurement involves visualization or quantitative methods, intending to assess and represent the model's uncertainty, providing a clearer understanding of the confidence associated with its predictions.
\subsection{Uncertainty Calibration}
Calibration can be classified into post-calibration and online calibration. Post-calibration refers to the process of re-mapping the predictions of a pre-trained model to yield more accurate probabilities. Common post-calibration techniques include Temperature Scaling, Vector Scaling, and others \cite{17guo2017}. In contrast, online calibration involves constraining predictive uncertainty during the model's training process, allowing the model to generate credible predictions directly.

We focus on multi-class tasks in visual fire detection, where the model takes a single image \(X\) as input, and the corresponding label is denoted as \(Y \in \{1, \ldots, K\}\), with \(K\) representing the number of classes. The model processes the input images and, after passing through the final softmax activation function, outputs the predicted probabilities for each class \(\{\hat{p}_1, \hat{p}_2, \ldots, \hat{p}_K\}\). The largest predicted probability is denoted as \(\hat{P}\), and the class corresponding to \(\hat{P}\) is the model's predicted class, denoted as \(\hat{Y}\). Therefore, \(\hat{P}\) represents the model's confidence for the sample belonging to class \(\hat{Y}\).

The ideal calibration result for a multi-class fire detection model is given by
\begin{equation}
\label{eq:1}
P(\hat{Y}=Y|\hat{P}=p) = p \quad\forall{p} \in [0,1]
\end{equation}
The equation implies that for all samples, the model's confidence in its predictions should be numerically consistent with its true classification accuracy. Specifically, if \( m \) samples all have a confidence level \( p \), then the model should correctly classify \( m \times p \) of these samples. The discrepancy between the model's predicted confidence and the actual observed accuracy across different confidence levels is what the calibration error measures. However, since \( \hat{P} \) in equation \ref{eq:1} is a continuous random variable, it is challenging to verify the equation's validity based on a finite number of samples. Therefore, binning or other approximation methods are typically employed to address this issue.

\subsection{Uncertainty Measurement}
Uncertainty is an abstract concept which must be assessed through either visual or quantitative methods. A reliability diagram is a visual method for reflecting the uncertainty of a model by representing the true classification accuracy as the function of confidence. The implementation steps are as follows: First, test samples are input into the trained model to obtain corresponding predicted probabilities and classification predictions. Next, the interval \( [0,1] \) is divided into \( M \) sub-intervals, and the predicted probabilities \( \hat{P} \) are assigned to one of the \( M \) intervals. Let \( B_{m} \) represents the set of samples whose predicted probabilities fall within the interval \( I_{m} = (\dfrac{m-1}{M},\dfrac{m}{M})\). The predicted accuracy for all samples in the \textit{m}-th interval can be expressed as
\begin{equation}
\label{eq:2}
acc(B_{m}) = \dfrac{\sum_{\substack{i\in B_{m}}} 1(\hat{y}_i=y_{i})}{|B_{m}|}
\end{equation}
where \(\hat{y_i}\) represents the predicted category of sample \( i \), \(y_{i}\) represents the true category of sample \( i \), \( B_{m} \) represents the set of samples falling in the \(m\)-th interval, and \(acc(B_{m})\) can be seen as an unbiased estimate of \(P(\hat{Y}=Y|\hat{P}\in I_{m})\).
The average confidence of all samples in the \(m\)-th sub-interval can be expressed as
\begin{equation}
\label{eq:3}
conf(B_{m}) = \dfrac{\sum_{\substack{i\in B_{m}}} \hat{p}_i}{|B_{m}|}
\end{equation}
where \(\hat{p}_i\) represents the predicted probability, also known as the confidence, of sample \(i\), and \(conf(B_m)\) can be considered as an approximation of the value of \(p\) on the right side of equation \ref{eq:1}.

\begin{figure}[!htbp]
    \centering
    \includegraphics[width=2.2in]{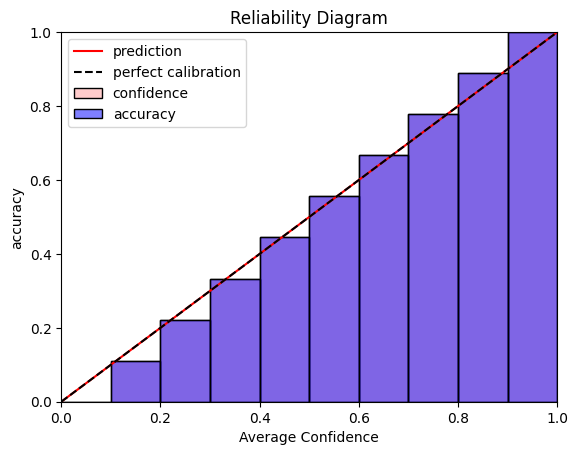}
    \caption{Reliability diagram under perfect calibration.}
    \label{fig:2}
\end{figure}

Therefore, equation \ref{eq:1} can be approximated as \(acc(B_m) = conf(B_m)\), meaning that, in the case of ideal calibration, the reliability diagram should display as the identity function. Taking \(M=10\), the reliability diagram corresponding to the ideal calibration is shown in Figure \ref{fig:2}. The closer the curve~(in red) is to the diagonal line~(in black), the better the calibration performance. In the diagram, the height of the pink bars represents the average confidence of samples in each sub-interval, while the height of the purple bars reflects the classification accuracy of samples in the corresponding sub-interval. Ideally, the two areas coincide completely.

\begin{figure}[!htbp]
    \centering
    \subfloat[]{\includegraphics[width=1.7in]{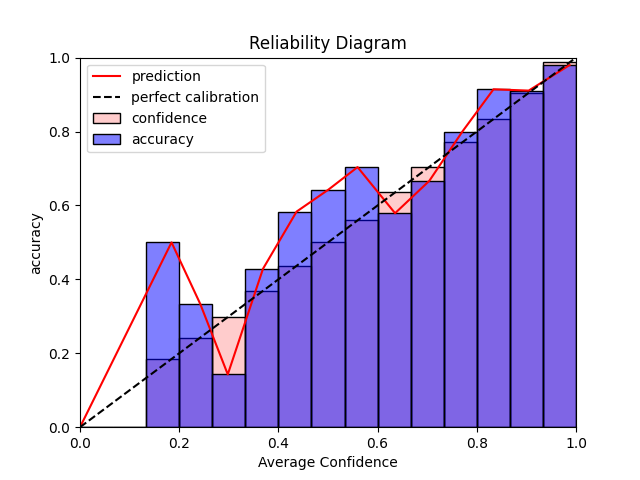}}\hfill
    \subfloat[]{\includegraphics[width=1.7in]{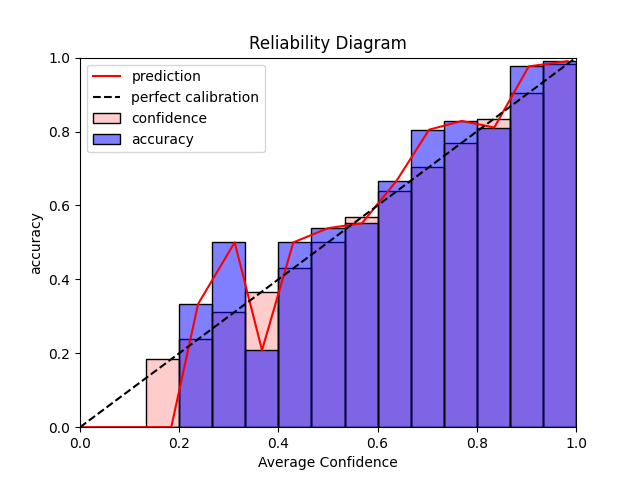}}
    \caption{Two indistinguishable reliability diagrams.}
    \label{fig:3}
\end{figure}

Reliability diagrams provide an intuitive means of reflecting a model's uncertainty. However, when the differences between two reliability diagrams are subtle, as illustrated in Figure \ref{fig:3}, it is hard to assess which model can provide more reliable decisions.

In such cases, representing the model's uncertainty using a scalar value proves to be more practical. A widely adopted method for quantification is illustrated as
\begin{equation}
\label{eq:4}
\sum_{\substack{\hat{p}}}|P(\hat{Y}=Y|\hat{P}=p) - p|
\end{equation}

This method of uncertainty quantification is derived from equation \ref{eq:1}, which assesses the predictive uncertainty by evaluating the gap between the model's confidence in its predictions and the true classification accuracy. This gap is typically approximated using the ECE\cite{20naeini2015} metric. The calculation of ECE follows a process similar to the construction of reliability diagrams: the interval \( [0,1] \) is divided into \(M\) sub-intervals, and within each sub-interval, the weighted average of the differences between the average accuracy and confidence is computed, as outlined in equation \ref{eq:5}, where \( n \) denotes the total number of samples. ECE evaluates the discrepancy between the predicted confidence and the true classification accuracy. A smaller ECE indicates better calibration performance of the model.
\begin{equation}
\label{eq:5}
ECE = \sum_{m=1}^{M} \dfrac{B_m}{n}|acc(B_m)-conf(B_m)|
\end{equation}

\subsection{ECE Loss}
We consider the approach of online calibration in visual fire detection, which involves constraining the model's credibility during training. One possible method is to use ECE as a loss function. However, the calculation of the accuracy \(\textit{acc}(B_m) \) in the ECE metric involves the 0-1 indicator function, as described in equation \ref{eq:2}. Therefore, ECE is non-differentiable, making it unsuitable for direct use as a loss function during optimization.

\begin{figure}[!htbp]
    \centering
    \subfloat{\includegraphics[width=2.8in]{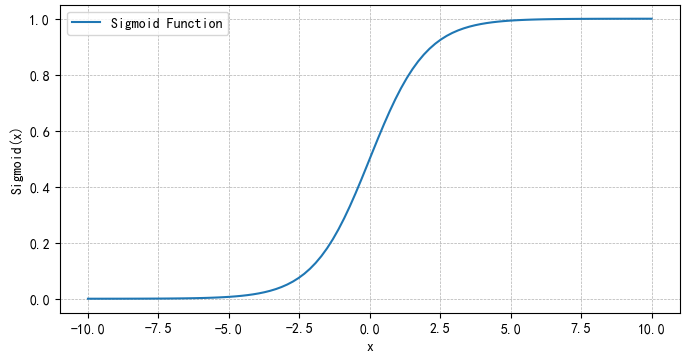}}
    \caption{Curve of sigmoid function.}
    \label{fig:4}
\end{figure}

The sigmoid function, with a range of (0, 1), can map any real number to this interval and exhibits a monotonically increasing behavior, facilitating a smooth transition between 0 and 1. Consequently, we propose approximating the indicator function with the sigmoid function. After this adjustment, the accuracy calculation is modified as presented in equation \ref{eq:6}. This converts the ECE metric into a differentiable ECE Loss, without altering the underlying calculation logic.
\begin{equation}
\label{eq:6}
acc(B_m)=\frac{\sum_{i \in B_m} S(tan(\pi \hat{p}_i-\dfrac{\pi}{2}))}{|B_m|}
\end{equation}
where the sigmoid function \textit{S(x)} is given in equation \ref{eq:7}, and its corresponding curve is depicted in Figure \ref{fig:4}.
\begin{equation}
\label{eq:7}
S(x)=\dfrac{1}{1+e^{-x}}
\end{equation}

\begin{figure}[!htbp]
    \centering
    \includegraphics[width=2.8in]{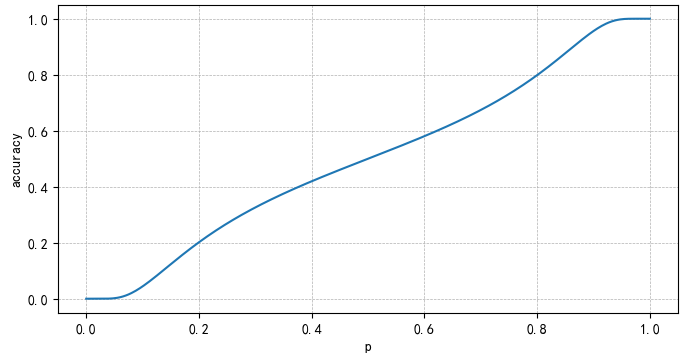}
    \caption{Curve of accuracy with predicted probability changing from zero to one.}
    \label{fig:5}
\end{figure}

When \(\hat{p}_i\) varies from 0 to 1, the curve of \(\textit{acc}(B_m)\) exhibits a smooth and differentiable profile, as depicted in Figure \ref{fig:5}.

\subsection{Online Calibration}
In this study, we combine ECE Loss with cross-entropy loss (NLL Loss) to jointly supervise the training of a multi-class fire detection model. Initially, we observe the relative magnitudes of NLL Loss and ECE Loss during training. Based on their proportions, we set the expected weight \(\gamma_E\) for ECE Loss, while fixing the weight coefficient of NLL Loss at 1.0. At the early stage of training, the model's classification performance is not stable, and thus the constraining effect of ECE Loss on model uncertainty is relatively weak. To balance the model's classification accuracy and reliable decision, we draw on the curriculum learning \cite{21bengio2009}. Specifically, we gradually increase the weight of ECE Loss as the number of training epochs progresses, until it reaches the predefined value of \(\gamma_E\). This strategy ensures that as the model becomes more confident and its accuracy improves, ECE Loss progressively contributes more to reducing uncertainty.

For DFAN model, the ratio of NLL Loss to ECE Loss is approximately 1:20, which leads to the choice of \(\gamma_E = 0.05\); for EdgeFireSmoke model, the ratio of NLL Loss to ECE Loss is approximately 5:1, so \(\gamma_E = 5\). The overall loss function used during model training can be expressed by
\begin{equation}
\label{eq:8}
L = L_n + \frac{c_e-s_e}{N-s_e}\times\gamma_E \times L_e
\end{equation}
where \( L_n \) represents the NLL Loss, and \( L_e \) the ECE Loss. When calculating ECE Loss, the number of sub-intervals \( M \) is set to 10. The variable \( c_e \) denotes the current training epoch, \( s_e \) denotes the epoch at which ECE Loss is first incorporated into the loss function, and \( N \) is the total number of epochs the model is trained for. Based on observations from the experimental process, as the training epoch increases, the model's classification accuracy gradually stabilizes and the weight of ECE Loss gradually increases, which ultimately approaches the expected weight. As a consequence, the constraining effect of ECE Loss on the model's uncertainty becomes more pronounced, encouraging the model to refine its uncertainty estimation and improve its overall calibration.

\section{Experiments and Analysis}
To evaluate the effectiveness of the proposed uncertainty modeling method, we conduct experiments using two publicly available multi-class datasets in the field of visual fire detection: DFAN and EdgeFireSmoke. The experimental procedure involves three main steps: First, the models are trained using only NLL Loss to establish a baseline. Next, the models are retrained by incorporating both NLL Loss and ECE Loss for comparison. Finally, we compare the classification performance and calibration performance of the models under the two settings to assess the improvements brought by ECE Loss.

\subsection{Dataset}
\begin{figure}
    \centering
    \subfloat[Boat Fire]{\includegraphics[width=1.6in, height=1.2in]{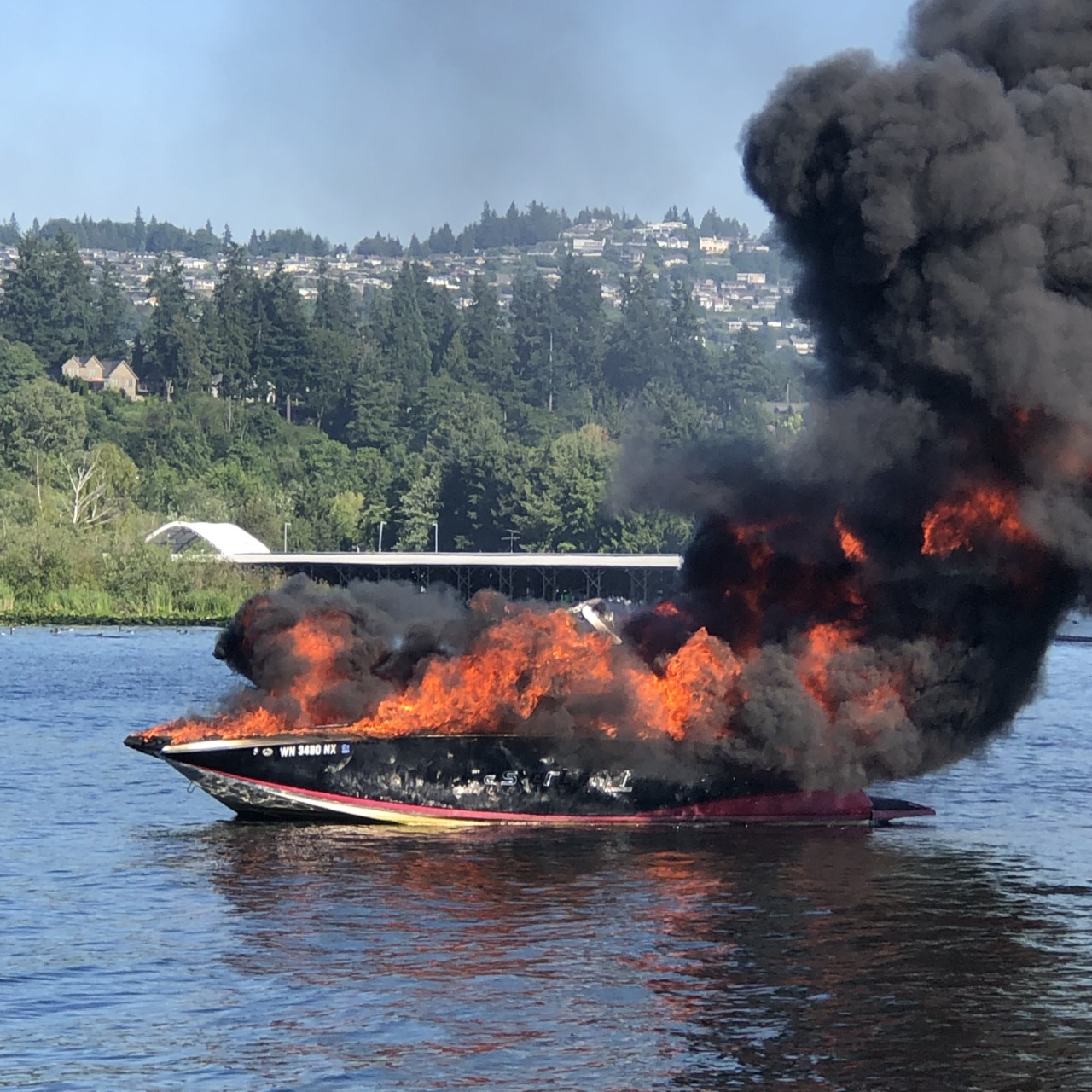}}\hfill
    \subfloat[Electric Fire]{\includegraphics[width=1.6in, height=1.2in]{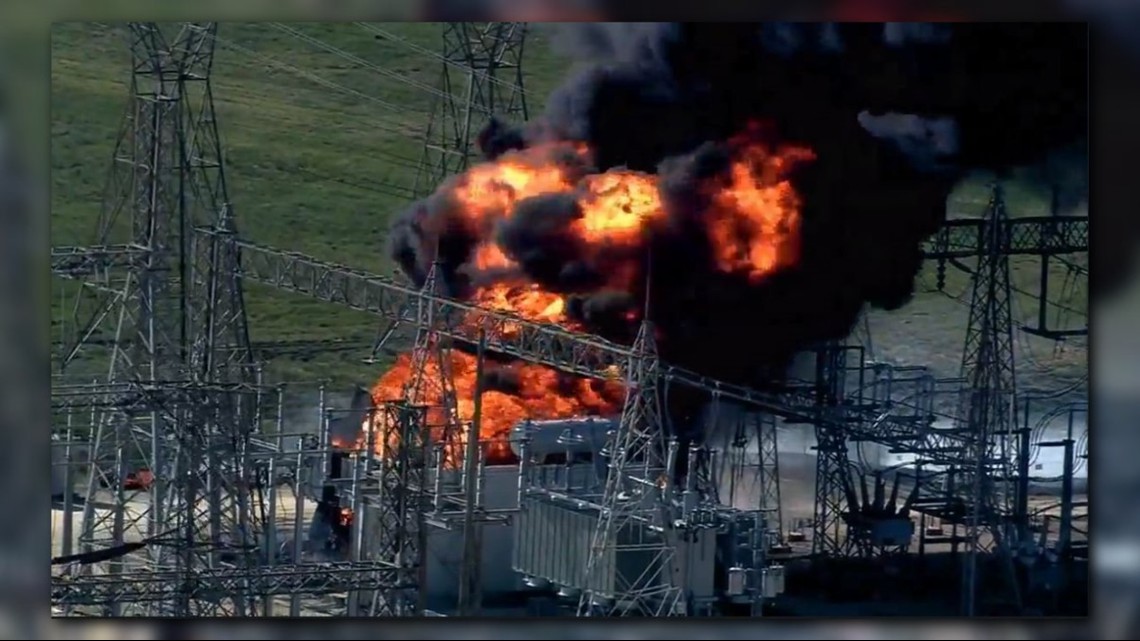}}\\
    \subfloat[Forest Fire]{\includegraphics[width=1.6in, height=1.2in]{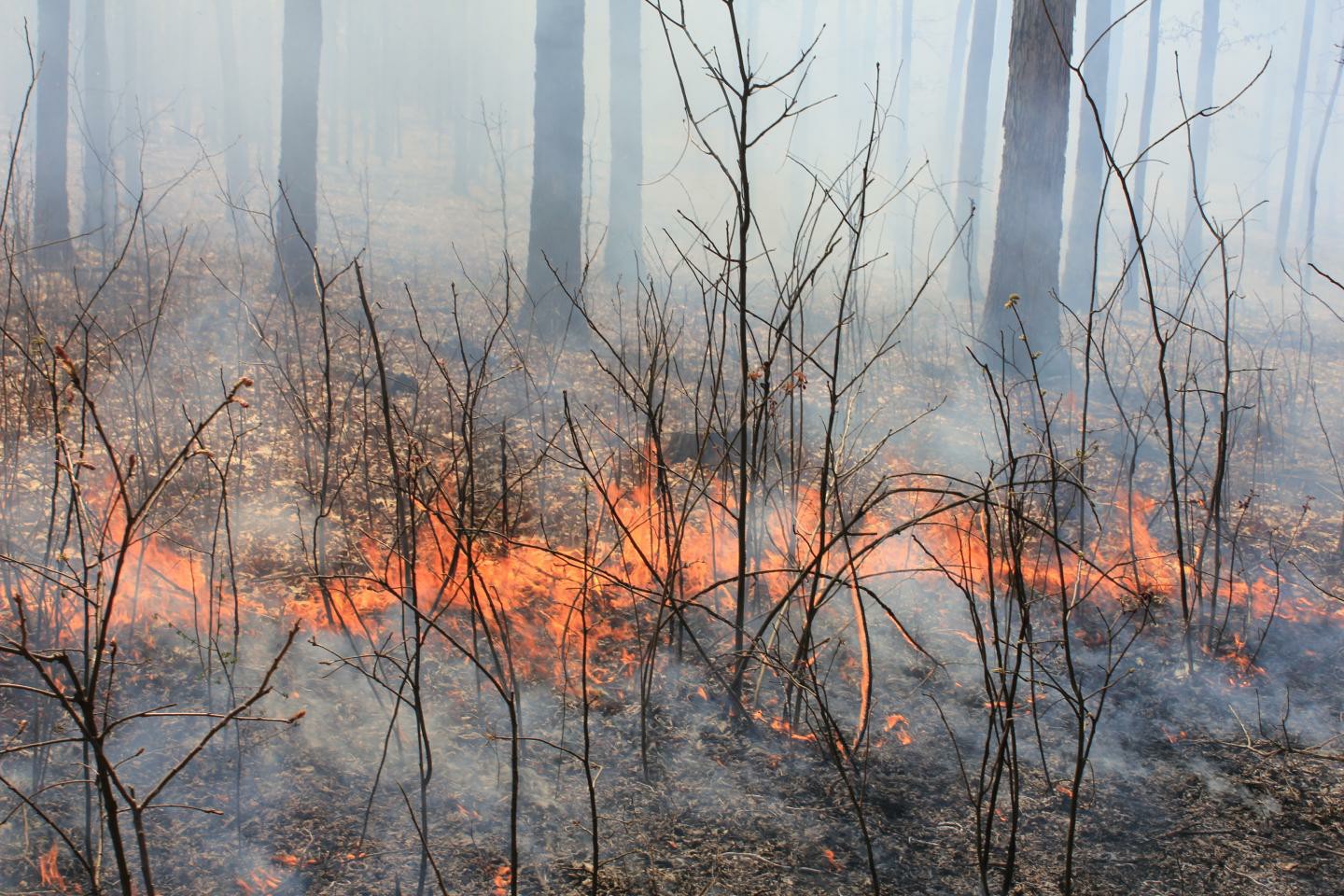}}\hfill
    \subfloat[Non Fire]{\includegraphics[width=1.6in, height=1.2in]{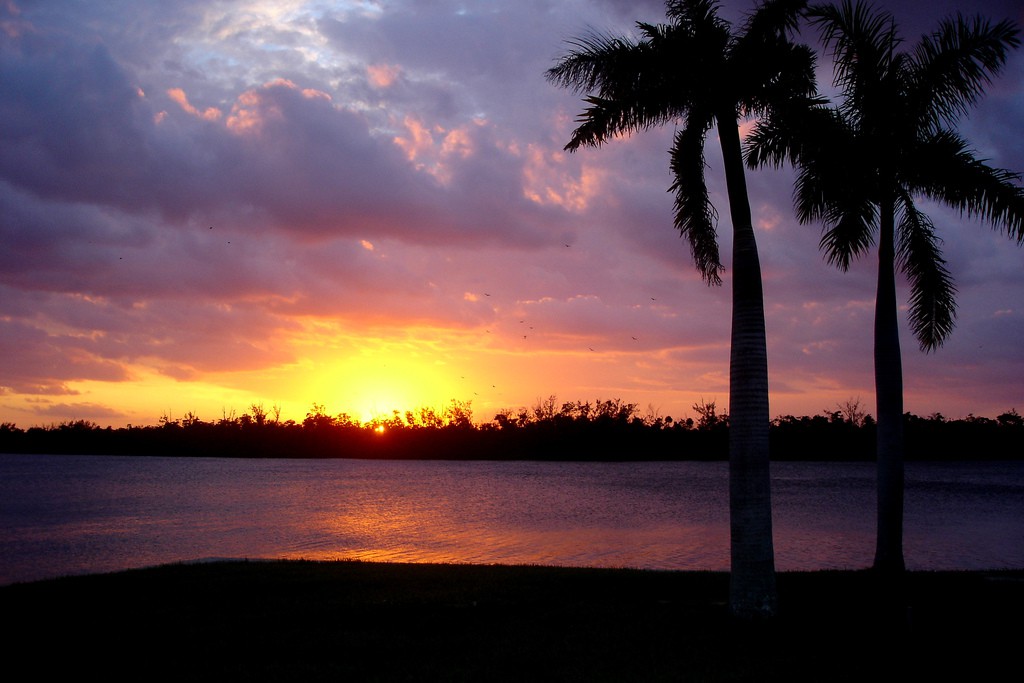}}\\
    \caption{Example images of the DFAN dataset.}
    \label{fig:6}
\end{figure}

The DFAN dataset \cite{22yar2022} is sourced from videos on platforms such as YouTube, Facebook, and disaster emergency management agencies, comprising a total of 3,803 images spread across twelve categories. The dataset is split into training, validation, and testing sets in a ratio of 7:2:1. The distribution of images across the different categories is presented in Table \ref{tab:table1}, and some example images from each category are shown in Figure \ref{fig:6}.

\begin{table}[!htbp]
\caption{\label{tab:table1}Class distribution of the DFAN dataset.}
\small
\centering
\renewcommand{\arraystretch}{1.25}
\begin{tabular}{cc}
\toprule
Category & Number of Pictures\\
\midrule
Boat Fire & 338\\
Building Fire & 305\\
Bus Fire & 400\\
Car Fire & 579\\
Cargo Fire & 207\\
Electric Fire & 300\\
Forest Fire & 480\\
Pickup Fire & 257\\
SUV Fire & 240\\
Van Fire & 300\\
Train Fire & 300\\
Non Fire & 97\\
\bottomrule
\end{tabular}
\end{table}

\begin{figure}
    \centering
    \subfloat[Burned-area]{\includegraphics[width=1.6in, height=1.2in]{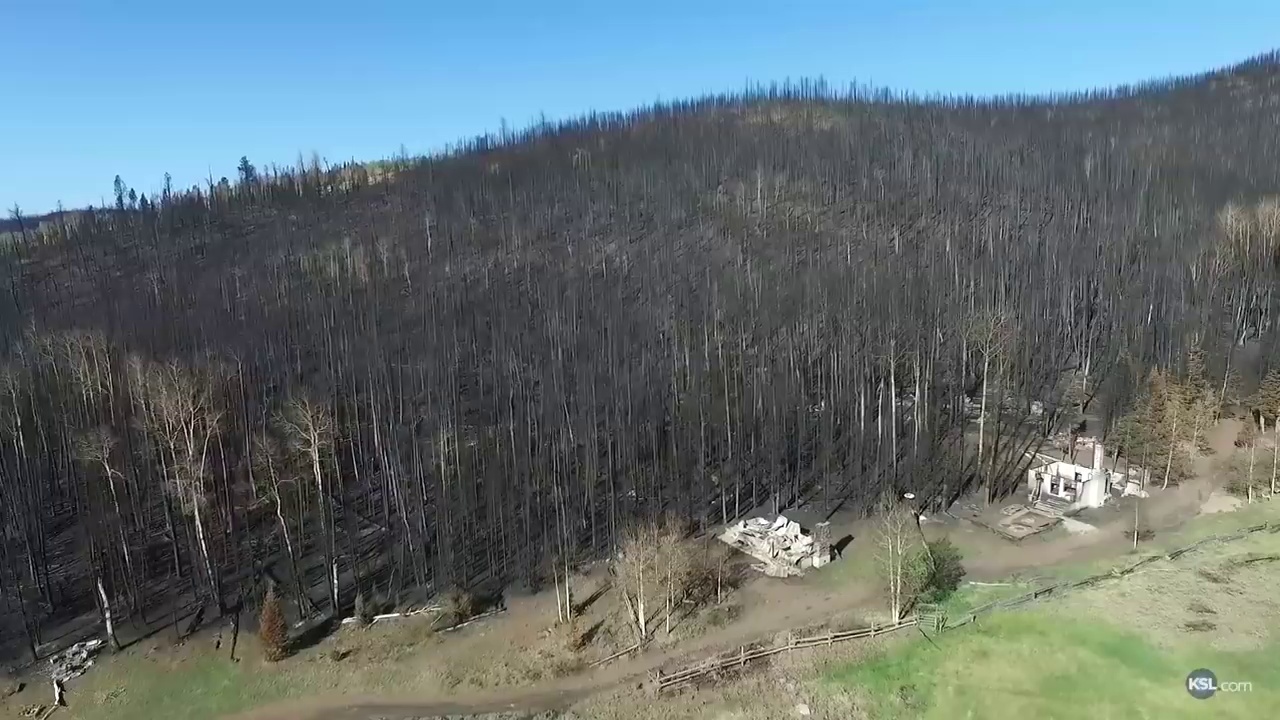}}\hfill
    \subfloat[Fire-smoke]{\includegraphics[width=1.6in, height=1.2in]{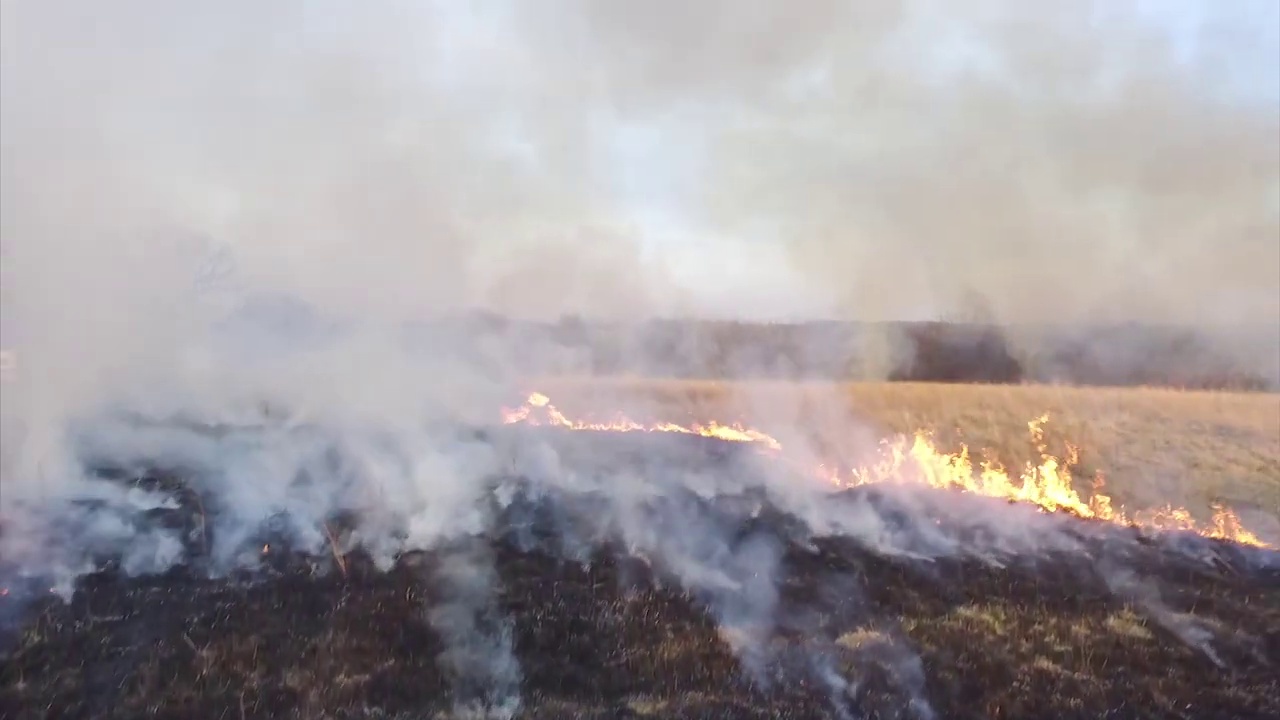}}\\
    \subfloat[Fog-area]{\includegraphics[width=1.6in, height=1.2in]{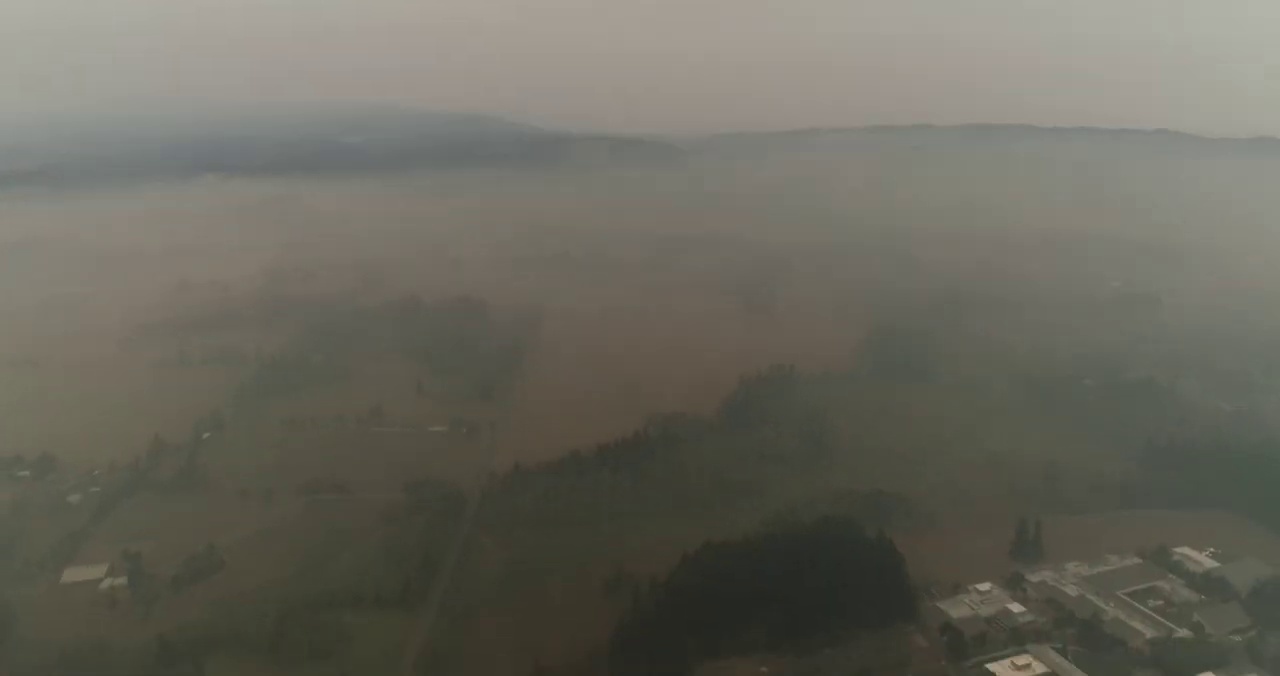}}\hfill
    \subfloat[Green-area]{\includegraphics[width=1.6in, height=1.2in]{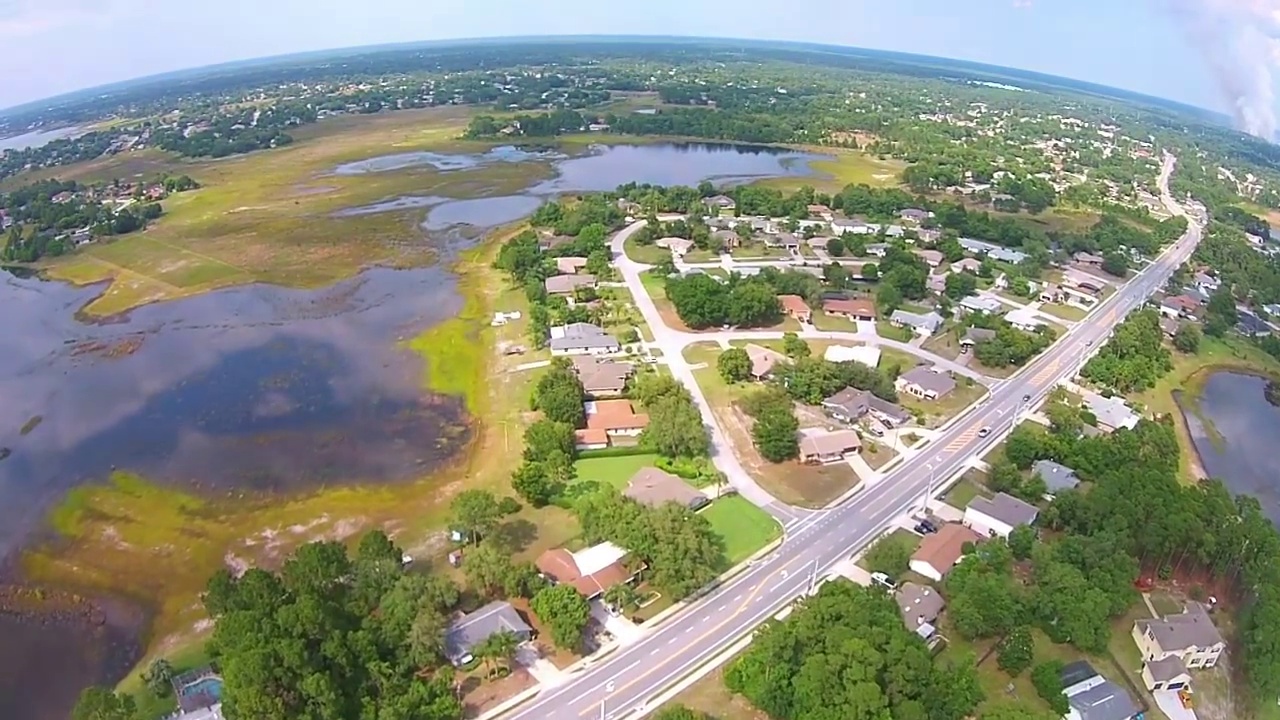}}\\
    \caption{Example images of EdgeFireSmoke dataset.}
    \label{fig:7}
\end{figure}

The EdgeFireSmoke dataset \cite{23almeida2022} consists of wildfire images captured by drones and is organized into four categories: Burned-area images, which typically feature blackened ground or withered tree trunks; Fire-smoke images, where the smoke is depicted as black or white; Fog-area images, characterized by blurred visuals and difficulty in distinguishing objects within the environment; and Green-area images, representing normal environments without the presence of the conditions mentioned above. The dataset contains a total of 49,452 images, with the data split into training, validation, and testing sets in a 2:3:5 ratio. The distribution of images across the different categories is shown in Table \ref{tab:table2}, and some example images from each category are illustrated in Figure \ref{fig:7}.

\begin{table}[!htbp]
\caption{\label{tab:table2}Class distribution of the EdgeFireSmoke dataset.}
\small
\centering
\renewcommand{\arraystretch}{1.25}
\begin{tabular}{cc}
\toprule
Category & Number of Pictures\\
\midrule
Burned-area & 9348\\
Fire-smoke & 15579\\
Fog-area & 9762\\
Green-area & 14763\\
\bottomrule
\end{tabular}
\end{table}

\subsection{Evaluation Metrics}
We employ five evaluation metrics to assess the performance of the multi-class fire detection models: precision, recall, F1 score, accuracy, and the ECE metric. The first four metrics are designed to evaluate the model's classification performance, while the ECE metric is intended to measure the decisions' reliability.

In the multi-classification task of this paper, precision, recall, F1 score, and accuracy are first calculated for each individual class, and then the average values across all classes are taken as the overall metrics. Besides, to measure the model's predictive uncertainty, the ECE metric is employed. The calculation of accuracy follows the method outlined in equation \ref{eq:2}. When plotting the reliability diagram and calculating ECE on the test data, the number of bins \(M\) is set to 15 to provide a detailed evaluation of the model's calibration across different confidence levels.

\subsection{Experimental Setup}
DFAN and EdgeFireSmoke models are implemented using the TensorFlow framework, and the training procedures largely adhere to the parameter settings outlined in the original papers. For the DFAN model, the training process is conducted over 50 epochs. To ensure the effective calibration of ECE Loss, and considering the relatively small size of the DFAN dataset, the batch size is increased to 32. Additionally, the input images are resized to 299×299 pixels. The model is optimized using the SGD optimizer with a learning rate set to 0.001. For the EdgeFireSmoke model, the training is performed over 30 epochs. To fully exploit the calibration effect of ECE Loss, the batch size is set to 128, and the images are resized to 224×224 pixels. The Adam optimizer is used with a learning rate of 0.001.

\subsection{Experimental Results}
\begin{figure}[!htbp]
    \centering
    \subfloat[\( \textit{DFAN}_{\textit{nll}} \)]{\includegraphics[width=1.7in, height=1.4in]{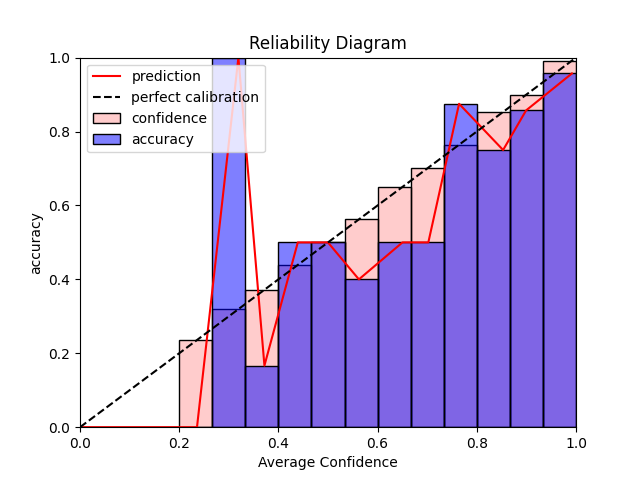}}\hfill
    \subfloat[\( \textit{DFAN}_{\textit{cali}} \)]{\includegraphics[width=1.7in, height=1.4in]{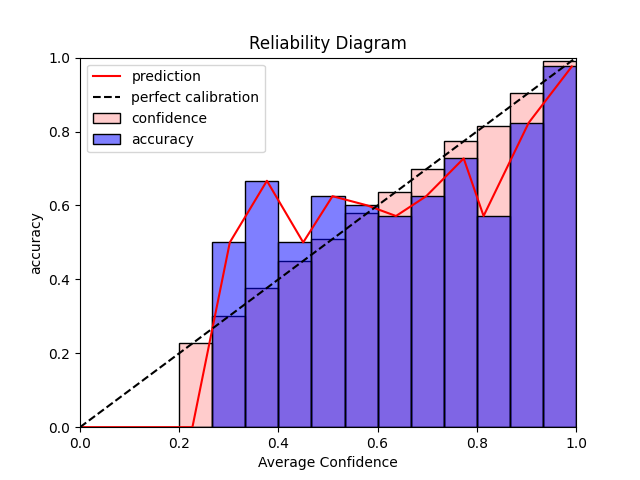}}
    \caption{Reliability diagram of DFAN model before and after calibration.}
    \label{fig:8}
\end{figure}

The model trained with NLL Loss is referred to as the vanilla model, while the model trained by combining NLL Loss and ECE Loss is referred to as the calibrated model. The vanilla DFAN model is denoted as \( \textit{DFAN}_{\textit{nll}} \), and the calibrated DFAN model is denoted as \(\textit{DFAN}_{\textit{cali}} \). Experiments results have demonstrated that the best calibration effect is achieved when \( s_e = 0 \), meaning that ECE Loss is incorporated from the start of training. The experimental comparison between the vanilla and calibrated DFAN models is shown in Table \ref{tab:table3}. Compared to the \( \textit{DFAN}_{\textit{nll}} \) model, the \(\textit{DFAN}_{\textit{cali}} \) model exhibits a significant reduction in uncertainty, with only a minor decrease of less than 0.7\% in classification accuracy. The reliability diagrams of the vanilla and calibrated models are shown in Figure \ref{fig:8}. The left diagram represents the reliability diagram for model \( \textit{DFAN}_{\textit{nll}} \). In the confidence intervals [0.2, 0.4] and [0.6, 0.8], a noticeable gap exists between the model's predicted confidence and its actual prediction accuracy. The function curve (the red curve in the diagram) deviates significantly from the perfect prediction (represented by the black dashed line), indicating a higher level of model uncertainty. In contrast, the right diagram shows the reliability diagram for model \(\textit{DFAN}_{\textit{cali}} \). In the interval [0.6, 0.8], the model's confidence and prediction accuracy are more closely aligned, with the function curve closely following the diagonal line. In the interval [0.2, 0.4], the prediction results have also improved, indicating better calibration. The calibrated model provides more reliable predictions, with better alignment between predicted confidence and classification accuracy.

\begin{table}[!htbp]
\caption{\label{tab:table3}Performance comparison of DFAN model before and after calibration.}
\small
\centering
\renewcommand{\arraystretch}{1.25}
\begin{tabular}{cccccc}
\toprule
Model & P(\%) & R(\%) & F1(\%) & ACC(\%) & ECE\\
\midrule
\( \textit{DFAN}_{\textit{nll}} \) & \textbf{87.94} & \textbf{87.54} & \textbf{87.31} & \textbf{87.54} & 0.05436\\
\( \textit{DFAN}_{\textit{cali}} \) & 87.72 & 86.89 & 86.75 & 86.89 & \textbf{0.04013}\\
\bottomrule
\end{tabular}
\end{table}

\begin{figure}[!htbp]
    \centering
    \subfloat[\( \textit{Edge}_{\textit{nll}} \)]{\includegraphics[width=1.7in, height=1.4in]{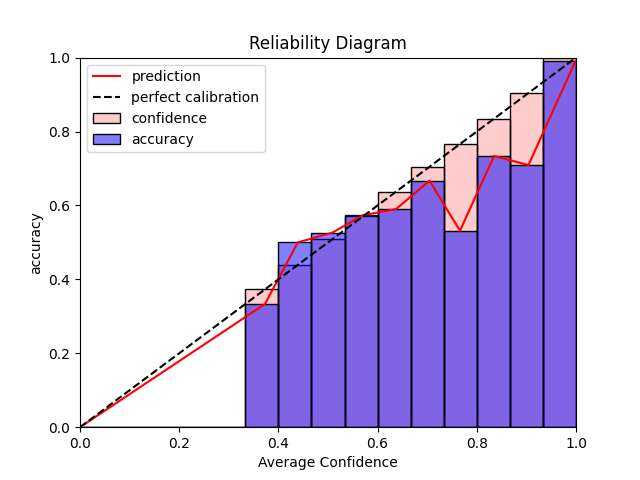}}\\
    \subfloat[\( \textit{Edge}_{\textit{cali1}} \)]{\includegraphics[width=1.7in, height=1.4in]{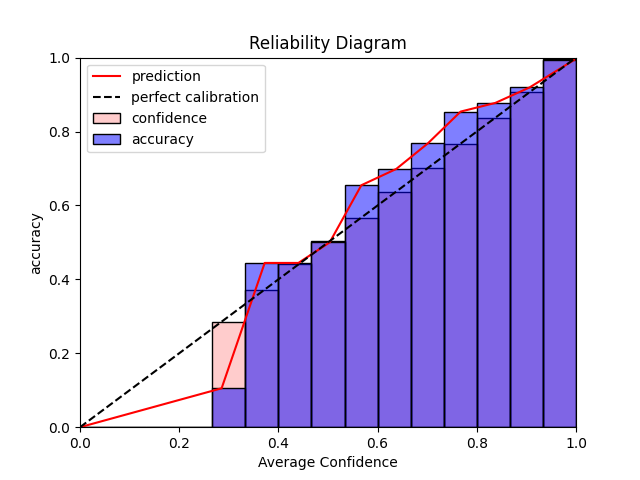}}\hfill
    \subfloat[\( \textit{Edge}_{\textit{cali2}} \)]{\includegraphics[width=1.7in, height=1.4in]{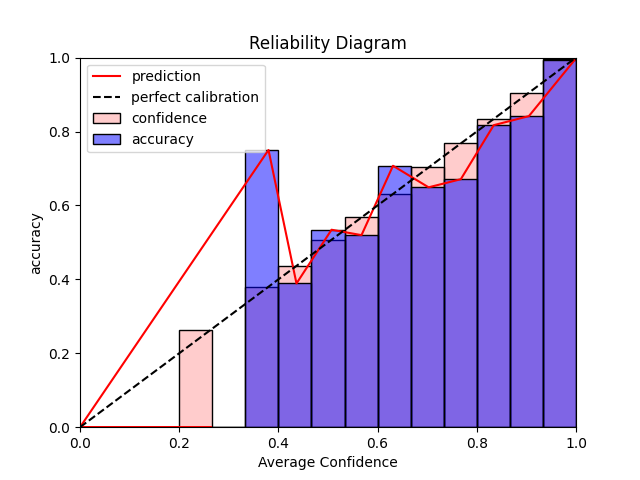}}\\
    \caption{Reliability diagram of EdgeFireSmoke model before and after calibration.}
    \label{fig:9}
\end{figure}

The experimental comparison between the vanilla and calibrated EdgeFireSmoke models is shown in Table \ref{tab:table4}. The vanilla EdgeFireSmoke model is denoted as \( \textit{Edge}_{\textit{nll}} \). In the online calibration experiments for this model, two experimental schemes demonstrate effective calibration effects. The first experimental scheme sets \( s_e = 0 \), and the resulting model is denoted as \( \textit{Edge}_{\textit{cali1}} \), corresponding to the second row of experimental results in Table \ref{tab:table4}; the second scheme sets \( s_e = 10 \), and the resulting model is denoted as \( \textit{Edge}_{\textit{cali2}} \), corresponding to the third row of experimental results in the table. Although the vanilla model already demonstrated low uncertainty, the introduction of ECE Loss still results in a decrease in model uncertainty, and the loss in model accuracy is controlled within 0.5\%. This indicates that the inclusion of ECE Loss successfully mitigates uncertainty without significantly sacrificing classification accuracy. The reliability diagrams of both the vanilla and calibrated EdgeFireSmoke models are shown in Figure \ref{fig:9}. In the interval [0.6, 1.0], the predicted accuracy and average confidence in both \( \textit{Edge}_{\textit{cali1}} \) and \(\textit{Edge}_{\textit{cali2}} \) models are much closer, highlighting the effectiveness of ECE Loss in reducing the model’s tendency towards \textit{overconfidence}. These results demonstrate that ECE Loss can improve the decisions' reliability by ensuring that the model's confidence aligns more accurately with its prediction accuracy.

\begin{table}[!htbp]
\caption{\label{tab:table4}Performance comparison of EdgeFireSmoke model before and after calibration.}
\small
\centering
\renewcommand{\arraystretch}{1.25}
\begin{tabular}{cccccc}
\toprule
Model & P(\%) & R(\%) & F1(\%) & ACC(\%) & ECE\\
\midrule
\( \textit{Edge}_{\textit{nll}} \) & \textbf{98.15} & \textbf{97.97} & \textbf{98.02} & \textbf{98.03} & 0.01208\\
\( \textit{Edge}_{\textit{cali1}} \) & 97.78 & 97.82 & 97.79 & 97.79 & \textbf{0.00596}\\
\( \textit{Edge}_{\textit{cali2}} \) & 97.72 & 97.95 & 97.80 & 97.80 & 0.00707\\
\bottomrule
\end{tabular}
\end{table}

Additionally, to balance the model's classification accuracy and reliable decision, a dynamic loss function design is proposed, where the weights of the loss functions change over time. This approach is inspired by the concept of curriculum learning, which allows the model to progressively incorporate more complex tasks as it stabilizes on simpler tasks. To verify the effectiveness of this approach, we consider incorporating ECE Loss from the beginning of the training process, with both loss functions' weights fixed at their expected values throughout training. This can be specifically represented by
\begin{equation}
\label{eq:9}
L = L_n + \gamma_E \times L_e
\end{equation}

\begin{figure}[!htbp]
    \centering
    \subfloat[\( \textit{Edge}_{\textit{cali1}} \)]{\includegraphics[width=1.7in, height=1.4in]{figs/EdgeFireSmoke_cali1.png}}\hfill
    \subfloat[\( \textit{Edge}_{\textit{w/o cl}} \)]{\includegraphics[width=1.7in, height=1.4in]{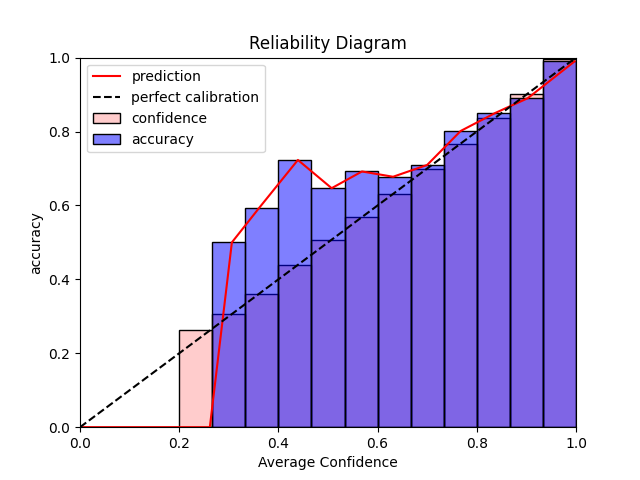}}\\
    \subfloat[\( \textit{DFAN}_{\textit{cali}} \)]{\includegraphics[width=1.7in, height=1.4in]{figs/DFAN_cali.png}}\hfill
    \subfloat[\( \textit{DFAN}_{\textit{w/o cl}} \)]{\includegraphics[width=1.7in, height=1.4in]{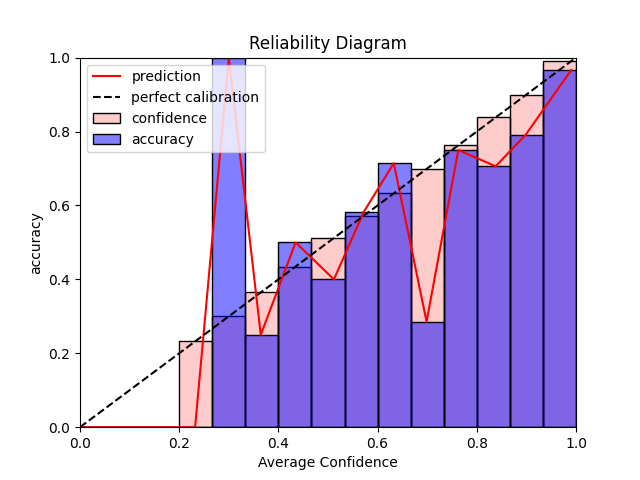}}\\
    \caption{Reliability diagram of EdgeFireSmoke model before and after calibration.}
    \label{fig:10}
\end{figure}

When training the model, the weight coefficient \(\gamma_E\) for ECE Loss remains constant, ensuring that the magnitudes of both loss functions are kept consistent and exert an equal constraining effect on the model's training process. Under this setting, the training results of the DFAN model are denoted as \( \textit{DFAN}_{\textit{w/o cl}} \), and those of the EdgeFireSmoke model are denoted as \( \textit{Edge}_{\textit{w/o cl}} \). The comparative experimental results are shown in Table \ref{tab:table5} and Figure \ref{fig:10}. The results demonstrate that, whether in terms of model classification accuracy or decision reliability, the models trained using curriculum learning approach outperforms the models with a constant weight setting for ECE Loss. This suggests that during the early stage of training, when the model's classification accuracy is still developing, introducing ECE Loss could even hinder the model's ability to learn effective classification patterns. Following the curriculum learning, the model first focus on the classification accuracy and later on the decision reliability. This gradual transition leads to more reliable and accurate predictions.

\begin{table}[!htbp]
\caption{\label{tab:table5}Performance comparison of models under different training settings.}
\small
\centering
\renewcommand{\arraystretch}{1.25}
\begin{tabular}{cccccc}
\toprule
Model & P(\%) & R(\%) & F1(\%) & ACC(\%) & ECE\\
\midrule
\( \textit{Edge}_{\textit{cali1}} \) & \textbf{97.78} & 97.82 & 97.79 & 97.79 & \textbf{0.00596}\\
\( \textit{Edge}_{\textit{cali2}} \) & 97.72 & \textbf{97.95} & \textbf{97.80} & \textbf{97.80} & 0.00707\\
\( \textit{Edge}_{\textit{w/o cl}} \) & 97.33 & 97.15 & 97.19 & 97.20 & 0.11289\\
\midrule
\( \textit{DFAN}_{\textit{cali}} \) & \textbf{87.72} & \textbf{86.89} & \textbf{86.75} & \textbf{86.89} & \textbf{0.04013}\\
\( \textit{DFAN}_{\textit{w/o cl}} \) & 86.61 & 85.90 & 85.55 & 85.90 & 0.05254\\
\bottomrule
\end{tabular}
\end{table}

\section{Conclusion}
To address the high predictive uncertainty in fire detection, we propose a new method to model uncertainty in visual fire detection by introducing a differentiable ECE Loss and calibrating the fire detection model online. Inspired by curriculum learning, we adjust the weight of ECE Loss over time, balancing the model's classification accuracy and reliable decision. Experiments conducted on two multi-class datasets, DFAN and EdgeFireSmoke, indicate that even when the predictive uncertainty is relatively low, the incorporation of ECE Loss can mitigate the model's tendency toward \textit{overconfidence}, effectively improving the decision reliability while keeping the sacrifice in classification accuracy within 0.7\%.

Given the limited availability of multi-class datasets in fire detection, we validate the effectiveness of our method with two commonly used datasets. The development of more classification datasets will further promote the adoption of uncertainty modeling techniques in visual fire detection. Future work will focus on improving the effectiveness of calibration, exploring methods to enhance classification accuracy while maintaining the reliability of decisions, and ultimately achieving better effects in real applications.

\bibliography{reference}

\end{document}